\documentclass[conference]{IEEEtran}
\IEEEoverridecommandlockouts                              

\usepackage{amsmath} 
\usepackage{amsthm}  
\usepackage{amsfonts}
\usepackage{amssymb}  
\usepackage{dsfont}
\usepackage{enumitem}
\usepackage[dvipsnames]{xcolor}
\usepackage{tikz}
\usepackage{pgfplots}
\usepackage[hidelinks]{hyperref}
\usepackage{orcidlink}
\pgfplotsset{compat=1.18,
	/pgfplots/ybar legend/.style={
		/pgfplots/legend image code/.code={%
			\draw[##1,/tikz/.cd,yshift=-0.25em]
			(0cm,0cm) rectangle (3pt,0.8em);},},
	tick label style={font=\small},
	label style={font=\small},
}
\usepackage[nolist,nohyperlinks]{acronym}
\usepackage[capitalize]{cleveref}
\Crefname{figure}{Fig.}{Figs.}
\crefname{equation}{}{}
\Crefname{equation}{Equation}{Equations}

\usepackage[per-mode=symbol,exponent-product=\cdot]{siunitx}

\usepackage{silence}  							
\usepackage[utf8]{inputenc}    
\usepackage[T1]{fontenc}       
\usepackage[style=ieee,doi=false,isbn=false,date=year,backend=biber,maxbibnames=15,maxcitenames=2,mincitenames=1,uniquelist=false,uniquename=false,giveninits=true]{biblatex}
\bibliography{bib}

\usepackage{booktabs}
\usepackage{tabularx}
\newcolumntype{Y}{>{\raggedright\arraybackslash}X}
\let\originalleft\left
\let\originalright\right
\renewcommand{\left}{\mathopen{}\mathclose\bgroup\originalleft}
\renewcommand{\right}{\aftergroup\egroup\originalright}

\newcommand{\bandwidth}{h}
\newcommand{\baseparameters}{z}
\newcommand{\cardinality}[1]{\mathrm{card}\left(#1\right)}
\newcommand{\density}[2]{p_{#1}\left(#2\right)}
\newcommand{\densityis}[2]{p_{#1}^*\left(#2\right)}
\newcommand{\densityfunc}[1]{p_{#1}}
\newcommand{\densityisfunc}[1]{p_{#1}^*}
\newcommand{\dimension}{d}
\newcommand{\expectation}[1]{\mathds{E}\left[ #1 \right]}
\newcommand{\indexscenario}{i}
\newcommand{\indexsimulation}{k}
\newcommand{\indexremaining}{j}
\newcommand{\kernelsymbol}{K}
\newcommand{\kernelfunc}[1]{\kernelsymbol\left(#1\right)}
\newcommand{\numberofcritical}{N_{\mathrm{C}}}
\newcommand{\numberofmc}{N_{\mathrm{MC}}}
\newcommand{\numberofis}{N_{\mathrm{NIS}}}
\newcommand{\numberofscenarios}{N}
\newcommand{\parameters}{x}
\newcommand{\realnumbers}{\mathds{R}}
\newcommand{\setindicestest}{I_{\mathrm{test}}}
\newcommand{\setindicaspareto}{I_{\mathrm{pareto}}}
\newcommand{\simulationoutcome}[1]{R\left(#1\right)}
\newcommand{\transformationsymbol}{f}
\newcommand{\transformation}{\transformationsymbol_{\transformationpararameters}}
\newcommand{\transformationinv}{\transformation^{-1}}
\newcommand{\transformationpararameters}{\theta}
\newcommand{\transformationfunc}[1]{\transformation\left(#1\right)}
\newcommand{\transformationfuncinv}[1]{\transformationinv\left(#1\right)}
\newcommand{\ud}{\mathrm{\,d}}

\begin{acronym}[AAAAAAAA]
	\acro{ADS}{Automated Driving System}\acroindefinite{ADS}{an}{an}
	\acro{FSM}{Fuzzy Safety Model}\acroindefinite{FSM}{an}{a}
	\acro{IQR}{Inter-Quartile Range}\acroindefinite{IQR}{an}{an}
	\acro{KDE}{Kernel Density Estimation}
	\acro{MAF}{Masked Autoregressive Flow}
	\acro{NF}{Normalizing Flows}
	\acro{PDF}{Probability Density Function}
\end{acronym}

\newlength{\figurewidth}
\newlength{\figureheight}
\newtheorem{researchquestion}{Research question}

\begin{document}

	\title{\LARGE \bf
		Comparing Normalizing Flows with Kernel Density Estimation in Estimating Risk of Automated Driving Systems%
		\thanks{The research presented in this work has been made possible by the SYNERGIES project.
			This project is funded by the European Union's Horizon Europe Research \& Innovation Actions under grant agreement No.\ 101146542. 
			Views and opinions expressed are however those of the authors only and do not necessarily reflect those of the European Union or the European Climate, Infrastructure and Environment Executive Agency (CINEA). 
			Neither the European Union nor the granting authority can be held responsible for them.}
	}
	
	\author{\IEEEauthorblockN{1\textsuperscript{st} Erwin de Gelder \orcidlink{0000-0003-4260-4294}} 
		\IEEEauthorblockA{\textit{TNO, Integrated Vehicle Safety} \\
			Helmond, the Netherlands}
			\and
			\IEEEauthorblockN{2\textsuperscript{nd} Maren Buermann \orcidlink{0009-0000-6740-8041}}
			\IEEEauthorblockA{\textit{TNO, Integrated Vehicle Safety} \\
			Helmond, the Netherlands }
			\and
			\IEEEauthorblockN{3\textsuperscript{rd} Olaf Op den Camp \orcidlink{0000-0002-6355-134X}}
			\IEEEauthorblockA{\textit{TNO, Integrated Vehicle Safety} \\
			Helmond, the Netherlands }
	}

	\maketitle
	\thispagestyle{empty}
	\pagestyle{empty}

	\begin{abstract}
	The development of safety validation methods is essential for the safe deployment and operation of \acp{ADS}. 
	One of the goals of safety validation is to prospectively evaluate the risk of \iac{ADS} dealing with real-world traffic. 
	Scenario-based assessment is a widely-used approach, where test cases are derived from real-world driving data.
	To allow for a quantitative analysis of the system performance, the exposure of the scenarios must be accurately estimated.
	The exposure of scenarios at parameter level is expressed using \iac{PDF}.
	However, assumptions about the \ac{PDF}, such as parameter independence, can introduce errors, while avoiding assumptions often leads to oversimplified models with limited parameters to mitigate the curse of dimensionality.
	
	This paper considers the use of \ac{NF} for estimating the \ac{PDF} of the parameters.
	\ac{NF} are a class of generative models that transform a simple base distribution into a complex one using a sequence of invertible and differentiable mappings, enabling flexible, high-dimensional density estimation without restrictive assumptions on the \ac{PDF}'s shape.
	We demonstrate the effectiveness of \ac{NF} in quantifying risk and risk uncertainty of \iac{ADS}, comparing its performance with \ac{KDE}, a traditional methods for non-parametric \ac{PDF} estimation.
	While \ac{NF} require more computational resources compared to \ac{KDE}, \ac{NF} is less sensitive to the curse of dimensionality.
	As a result, \ac{NF} can improve risk uncertainty estimation, offering a more precise assessment of \iac{ADS}'s safety. 
	
	This work illustrates the potential of \ac{NF} in scenario-based safety.
	Future work involves experimenting more with using \ac{NF} for scenario generation and optimizing the \ac{NF} architecture, transformation types, and training hyperparameters to further enhance their applicability.
\end{abstract}
    \acresetall
    
	\section{INTRODUCTION}
\label{sec:introduction}

\Acp{ADS} are expected to improve traffic safety, comfort, and congestion by reducing human errors \autocite{chan2017advancements}. 
While lower-level systems like adaptive cruise control and lane keeping assist are already common, higher-level \acp{ADS} (SAE Level 3 and 4) are nearing large-scale deployment. 
Ensuring their safety prior to release is critical. 
As large-scale road testing is impractical \autocite{kalra2016driving}, prospective methods are required.
Scenario-based safety validation has emerged as a widely supported approach in the automotive domain \autocite{riedmaier2020survey, degelder2024streetwise}.

An essential aspect of scenario-based safety validation is risk estimation, where risk is the combination of exposure to a scenario and the consequence of the response to that scenario.
\Iac{PDF} measures the exposure to different scenarios at the parameter level. 
However, making assumptions about the \ac{PDF}, e.g., assuming parameter independence, can introduce errors.
Conversely, avoiding such assumptions often results in oversimplified models that include only a limited number of parameters to address the curse of dimensionality.

\Ac{KDE} \autocite{rosenblatt1956remarks, parzen1962estimation} is a non-parametric method for estimating the \ac{PDF} that does not rely on assumptions regarding the shape of the \ac{PDF}. 
In the literature, \ac{KDE} has been utilized to estimate the \ac{PDF} of scenario parameters for the assessment of \acp{ADS} \autocite{degelder2021risk, degelder2023certain}. 
However, a notable disadvantage of \ac{KDE} is its susceptibility to the curse of dimensionality. 
As an alternative, this paper explores the use of \ac{NF} \autocite{papamakarios2021normalizing}, another non-parametric method for \ac{PDF} estimation, which has demonstrated promising results for high-dimensional data.

In this work, we will compare the use of \ac{KDE} and \ac{NF} in estimating the exposure of scenarios at the parameter level and the associated risk estimated using scenarios sampled from the estimated \acp{PDF}. 
To the best of our knowledge, \ac{NF} have not been utilized in the literature on safety assessment methods for \acp{ADS}.
Furthermore, several recent survey papers on scenario generation \autocite{batsch2020taxonomy, cai2022survey, schutt20231001, ding2023survey, cai2024review, kayisu2024comprehensive} do not mention the application of \ac{NF}.
Consequently, this study aims to provide an initial perspective on the potential of \ac{NF} for the safety assessment of \acp{ADS}.
Specifically, this work will address two different aspects of safety assessment.
The first part involves comparing the use of \ac{NF} and \ac{KDE} in estimating the \ac{PDF} of scenario parameters, leading to the following research question:
\begin{researchquestion}
	\label{rq:exposure}
	How do \ac{NF} and \ac{KDE} compare in estimating the \ac{PDF} of scenario parameters?
\end{researchquestion}
The second aspect focuses on comparing \ac{NF} and \ac{KDE} in estimating the risk, following the procedure outlined in \autocite{degelder2021risk}.
Thus, the second research question is:
\begin{researchquestion}
	\label{rq:risk}
	When used for scenario exposure modeling, how do \ac{NF} and \ac{KDE} compare in quantifying \ac{ADS} risk? 
\end{researchquestion}

\begin{figure*}
	\centering
	\definecolor{blockline}{RGB}{0,0,255}
	\definecolor{blockbg}{RGB}{204,204,255}
	\newlength{\blockwidth}\setlength{\blockwidth}{9.98em}
	\newlength{\blockwidthsmall}\setlength{\blockwidthsmall}{7.78em}
	\newlength{\blockheight}\setlength{\blockheight}{6.5em}
	\newlength{\blocksep}\setlength{\blocksep}{1.45em}
	\tikzstyle{block}=[draw=blockline, fill=blockbg, minimum width=\blockwidth, text width=\blockwidth-0.5em, minimum height=\blockheight, align=center, node distance=\blockwidth+\blocksep]
	\tikzstyle{blocksmall}=[block, minimum width=\blockwidthsmall, text width=\blockwidthsmall-0.5em]
	\tikzstyle{arrow}=[->, line width=1pt, draw=blockline]
	\begin{tikzpicture}
		\node[blocksmall](input){%
			\textbf{Input}\\ 
			PDF of $\parameters$: $\densityfunc{\parameters}$, estimated using KDE or NF
		};
		
		\node[block, right of=input, node distance=(\blockwidthsmall+\blockwidth)/2+\blocksep](mc){%
			\textbf{Monte Carlo}\\
			Sample $\numberofmc$ times from $\densityfunc{\parameters}$, 
			and do simulations to obtain 
			$\left\{
			\parameters_{\indexsimulation},
			\simulationoutcome{\parameters_{\indexsimulation}}
			\right\}_{\indexsimulation=1}^{\numberofmc}$
		};
		
		\node[block, right of=mc](id){%
			\textbf{Importance density}\\
			Select $\numberofcritical<\numberofmc$ most critical scenarios and construct $\densityisfunc{\parameters}$ using KDE
		};
		
		\node[block, right of=id](is){%
			\textbf{Importance sampling}\\
			Sample $\numberofis$ times from $\densityisfunc{\parameters}$, 
			and do simulations to obtain 
			$\left\{
			\parameters_{\indexsimulation},
			\simulationoutcome{\parameters_{\indexsimulation}}
			\right\}_{\indexsimulation=1}^{\numberofis}$
		};
		
		\node[blocksmall, right of=is, node distance=(\blockwidthsmall+\blockwidth)/2+\blocksep](outcome){%
			\textbf{Result}\\
			Estimate collision probability, $\expectation{\simulationoutcome{\parameters}}$, using (3).
		};
		
		\draw[arrow] (input) -- (mc);
		\draw[arrow] (mc) -- (id);
		\draw[arrow] (id) -- (is);
		\draw[arrow] (is) -- (outcome);
	\end{tikzpicture}
	\caption{Schematic overview of the risk quantification method.}
	\label{fig:rq approach}
\end{figure*}
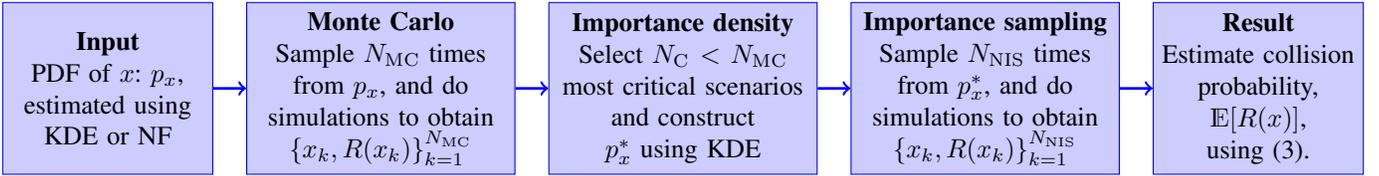

This work is structured as follows.
\Cref{sec:method} first briefly explains the risk quantification method of \autocite{degelder2021risk}, after which we explain how \ac{NF} and \ac{KDE} are exploited in this paper.
In \cref{sec:setup}, the case study is first detailed, after which the results aiming to answer the aforementioned research questions are presented.
This work concludes with a discussion in \cref{sec:discussion} and conclusions and a future outlook in \cref{sec:conclusions}.

	\section{METHOD}
\label{sec:method}

This section first clarifies how we quantify the risk of \iac{ADS}.
This also shows that \iac{PDF} is needed to estimate the exposure of scenarios at parameter level.
Since the shape of the \ac{PDF} is generally unknown beforehand, we propose to use a non-parametric method for estimating the \ac{PDF}.
This work employs \ac{NF} for non-parametric \ac{PDF} estimation and compares this with the more traditional \ac{KDE}.
\Cref{sec:nf} explains \ac{NF} and how this work utilizes them and \cref{sec:kde} provides further details on \ac{KDE}.

\subsection{Risk quantification}
\label{sec:risk quantification}

As explained in \autocite{degelder2021risk} and in line with the ISO~26262 standard \autocite{ISO26262},  risk consists of three components: exposure, severity, and controllability (definitions are in \autocite{degelder2021risk, ISO26262}).
When calculating severity, the extent of harm in a potential collision should be considered.
For the sake of comparing \ac{NF} with \ac{KDE}, however, this work treats all collisions equally.
Thus, we will express the risk as the probability of a collision given that the \ac{ADS} deals with a certain predefined type of scenario.
\Cref{fig:rq approach} summarizes the risk quantification approach that is used in this work.

Let $\simulationoutcome{\parameters}$ denote the simulation outcome of a scenario that is parameterized by $\parameters\in\realnumbers^{\dimension}$.
If $\simulationoutcome{\parameters}=1$ denotes a simulation run that ends in a crash and $\simulationoutcome{\parameters}=0$ otherwise, then $\expectation{\simulationoutcome{\parameters}}$ is the expected probability of a collision, which is what we aim to estimate.
To estimate this, we need to know how $\parameters$ is distributed. 
As the underlying \ac{PDF} of $\parameters$ is generally unknown, this needs to be estimated.
For this, we will use either \ac{NF} (\cref{sec:nf}) or \ac{KDE} (\cref{sec:kde}).
For now, we assume that the \ac{PDF} denoted by $\densityfunc{\parameters}$, is known and that sampling from the \ac{PDF} is possible.
A straightforward way to estimate $\expectation{\simulationoutcome{\parameters}}$ is using a crude Monte Carlo simulation:
\begin{align}
	\label{eq:risk}
	\expectation{\simulationoutcome{\parameters}}
	&= \int_{\realnumbers^{\dimension}} 
	\simulationoutcome{\parameters} \density{\parameters}{\parameters} \ud\parameters \\
	\label{eq:mc}
	&\approx \frac{1}{\numberofmc}
	\sum_{\indexsimulation=1}^{\numberofmc} \simulationoutcome{\parameters_{\indexsimulation}},	\quad\parameters_{\indexsimulation}\sim\densityfunc{\parameters},
\end{align}
where $\numberofmc$ denotes the number of simulation runs.

In practice, the expected probability of a collision is low.
As a result, many simulations are required to obtain enough confidence in \cref{eq:mc}.
To accelerate the evaluation, importance sampling is utilized \autocite{degelder2021risk}. 
With importance sampling, we sample from the importance density, $\densityisfunc{\parameters}$, instead of sampling from $\densityfunc{\parameters}$. 
To acquire an unbiased estimate of $\expectation{\simulationoutcome{\parameters}}$, the simulation results are weighted to correct for the fact that we sample from $\densityisfunc{\parameters}$ instead of $\densityfunc{\parameters}$:
\begin{equation}
	\label{eq:is}
	\expectation{\simulationoutcome{\parameters}}
	\approx \frac{1}{\numberofis} 
	\sum_{\indexsimulation=1}^{\numberofis} \simulationoutcome{\parameters_{\indexsimulation}}
	\frac{ \density{\parameters}{\parameters_{\indexsimulation}} }{ \densityis{\parameters}{\parameters_{\indexsimulation}} },
	\quad\parameters_{\indexsimulation}\sim\densityisfunc{\parameters},
\end{equation}
where $\numberofis$ denotes the number of simulation runs with importance sampling.

Ideally, $\densityisfunc{\parameters}$ is chosen such that the estimation error of \cref{eq:is} is minimized.
However, this requires the value of $\expectation{\simulationoutcome{\parameters}}$ and a functional form of $\simulationoutcome{\parameters}$; both are unavailable. 
Therefore, we will follow the approach in \autocite{degelder2021risk} and first conduct the crude Monte Carlo sampling and determine $\densityisfunc{\parameters}$ on the results thereof.
More specifically, we select the $\numberofcritical$ most critical scenarios, where the criticality of a simulated scenario is based on the minimum time to collision.
Since the $\numberofcritical$ most critical scenarios is a subset of the $\numberofmc$ scenarios, $\numberofcritical < \numberofmc$.
Note that the time-to-collision is an appropriate metric for the scenario considered in our case study \autocite{mullakkal2017comparative} (\cref{sec:case study}), but this might not be an appropriate metric to measure the criticality in simulations of other type of scenarios.
Based on those $\numberofcritical$ scenarios, $\densityisfunc{\parameters}$ is constructed with \ac{KDE}, which is explained in \cref{sec:kde}.
\ac{KDE} is preferred over \ac{NF} for constructing the importance density due to its computational simplicity. 
Moreover, while \ac{KDE} may result in a less optimal approximation of $\expectation{\simulationoutcome{\parameters}}$ in terms of the uncertainty, we will demonstrate and discuss in the results that it exhibits a lower risk of introducing bias in the estimate of $\expectation{\simulationoutcome{\parameters}}$.

\subsection{Normalizing flows}
\label{sec:nf}

\Ac{NF} offer a flexible, deep-learning-based method capable of modeling complex, high-dimensional distributions. 
\ac{NF} are a class of generative models that estimate complex probability distributions by transforming a simple base distribution through a
sequence of invertible and differentiable mappings \autocite{rezende2015variational}. 

Where $\densityfunc{\parameters}$ denotes the unknown \ac{PDF} of $\parameters$, let  $\densityfunc{\baseparameters}:\realnumbers^{\dimension} \rightarrow \realnumbers$ denote a known and tractable density of a random variable with realization $\baseparameters\in\realnumbers^{\dimension}$.
A typical choice for $\densityfunc{\baseparameters}$ is the standard Gaussian distribution.
The main idea of \ac{NF} is to estimate a transformation function $\transformation:\realnumbers^{\dimension} \rightarrow \realnumbers^{\dimension}$, parameterized by $\transformationpararameters$, such that $\parameters=\transformationfunc{\baseparameters}$.
Now, we can compute the \ac{PDF} of $\parameters$ using the change of variables formula \autocite{papamakarios2021normalizing}:
\begin{equation}
	\label{eq:nf}
	\density{\parameters}{\parameters} 
	= \density{\baseparameters}{\transformationfuncinv{\parameters}}
	\left| \det \left( \frac{\partial\transformationfuncinv{\parameters}}{\partial\parameters} \right) \right|
\end{equation}
where the magnitude of the determinant of the Jacobian, $\left| \det \left( \frac{\partial\transformationfuncinv{\parameters}}{\partial\parameters} \right) \right|$, ensures that the requirements of a density are maintained after applying the transformation $\transformationinv$.
To ensure that \cref{eq:nf} is tractable, the transformation $\transformation$ must be easy to invert and its Jacobian must be easy to compute.
An important point, as noted in \autocite{papamakarios2017masked}, is that if two transformations $h$ and $g$ have these two properties, then so has their composition $g \circ h$.
Thus, the transformation $\transformation$ can be made deeper by composing multiple instances of it with different parameters $\theta$. 

In this work, the following implementation of \ac{NF} is adopted. 
A composition of four transformation functions is used, where each transformation function is a coupling layer, consisting of:
\begin{itemize}
	\item \Iac{MAF} layer \autocite{papamakarios2017masked}, where a residual network with two pre-activation residual blocks are used with each \ac{MAF} \autocite{he2016identity}. 
	We opted to use this type of transformer due to its simplicity and sufficiency for our purposes, making more complex alternatives unnecessary in this context.
	Each block uses two masked dense layers with $2\dimension$ hidden features each.
	During training, a dropout rate of \SI{20}{\percent} is used.
	
	\item A batch normalization layer to improve optimization \autocite{santurkar2018how}.
	
	\item A random permutation matrix fixed at the start of the training.
	This is necessary because the \ac{MAF} layer leaves part of the input unchanged. 
	That is also why at least three coupling layers are necessary, while at least four are preferred \autocite{dinh2014nice}.
\end{itemize}
As the base distribution, we use a standard $\dimension$-dimensional Gaussian distribution with zero mean and an identity matrix as variance.
During training, \SI{80}{\percent} of the available data is used to optimize the transformation parameters $\transformationpararameters$.
The remaining \SI{20}{\percent} is used to test how well the estimation performs.
We employed the Adam optimizer \autocite{kingma2015adam}, a widely-used, first-order, gradient-based optimization algorithm implemented in PyTorch.
A maximum of \num{5000} iterations are used during optimization, but if the performance is not improving over \num{100} iterations, the optimization is terminated. 
This whole procedure is repeated four times with different initializations of $\transformationpararameters$.
Then, $\transformationpararameters$ is based on the best performance during the different iterations.

\subsection{Kernel density estimation}
\label{sec:kde}

\ac{KDE} \autocite{rosenblatt1956remarks, parzen1962estimation} is widely used due to its simplicity and ability to approximate arbitrary distributions without assuming an underlying parametric model.
Given a set of $\numberofscenarios$ scenario parameters observed in data, $\{\parameters_{\indexscenario}\}_{\indexscenario=1}^{\numberofscenarios}$, \ac{KDE} estimates the density at $\parameters$ by averaging kernel functions $\kernelsymbol$ centered around each data point:
\begin{equation}
	\label{eq:kde}
	\density{\parameters}{\parameters} 
	= \frac{1}{\numberofscenarios\bandwidth^{\dimension}}
	\sum_{\indexscenario=1}^{\numberofscenarios} 
	\kernelfunc{ \frac{1}{\bandwidth} \left( \parameters - \parameters_\indexscenario \right) },
\end{equation}
where $\bandwidth$ is the bandwidth parameter that controls the level of smoothing.
The choice of $\bandwidth$ is crucial --- too small a bandwidth leads to overfitting, while too large a bandwidth oversmooths the \ac{PDF} estimate. 
A common method for determining $\bandwidth$ is leave-one-out cross-validation because this minimizes the difference between the real \ac{PDF} and the estimated \ac{PDF} according to the Kullback-Leibler divergence \autocite{turlach1993bandwidthselection}.
Commonly used kernel functions include Gaussian, Epanechnikov, and uniform kernels \autocite{silverman1986density}.
The choice of the Kernel function, $\kernelsymbol$, is commonly considered less important than the bandwidth. 

In this paper, we will adopt the often-used Gaussian kernel.
For the bandwidth selection, leave-one-out cross-validation is used because of the aforementioned reason.

	\section{CASE STUDY}
\label{sec:case study}

This section explains the case study that has been conducted in order to answer the research questions discussed in the introduction.
\Cref{sec:setup} explains how the case study has been conducted.
In \cref{sec:results rq1}, results related to \cref{rq:exposure} are discussed.
\Cref{sec:results rq2} aims to provide an answer to \cref{rq:risk}.

\subsection{Setup case study}
\label{sec:setup}

To demonstrate the use of \ac{NF} and \ac{KDE} for quantifying the risk, we will estimate the risk of a collision for a driver model based on the \ac{FSM} \autocite{mattas2022driver}.
We have chosen the \ac{FSM} because it has successfully been used to provide a benchmark for \acp{ADS} \autocite{mattas2022driver}. 
The \ac{FSM} uses two surrogate safety metrics for rear-end collision, the proactive fuzzy safety and critical fuzzy safety \autocite{mattas2020fuzzy}. 
The vehicle will start to gently brake as soon as the former is nonzero, while heavier braking will start as soon as the latter becomes nonzero.

For the risk quantification, this paper considers cut-in scenarios.
In these scenarios, another vehicle is driving in the lane next to the ego vehicle while initiating a lane change towards the lane of the ego vehicle.
After the lane change, the other vehicle becomes the leading vehicle of the ego vehicle.
The scenarios are parameterized using four parameters:
\begin{enumerate}
	\item the initial longitudinal velocity of the ego vehicle;
	\item the longitudinal velocity of the other vehicle, which is assumed to be constant throughout the scenario;
	\item the initial lateral velocity of the other vehicle, which is assumed to be constant until the vehicle completed its lane change, after which the lateral velocity becomes zero; and
	\item the distance between the ego vehicle and the other vehicle at the start of the scenario.
\end{enumerate}

To obtain realistic parameter values, cut-in scenarios are extracted from the HighD dataset \autocite{krajewski2018highD}. 
The approach presented in \autocite{degelder2020scenariomining} has been used to extract scenarios from a dataset, while \autocite{degelder2024parameterization} provide more specific details on how scenarios are extracted from the HighD dataset.
In total, \num{2916} cut-in scenarios are collected for this work's case study.

To study the influence of the amount of data, the experiments are repeated for various amount of data.
Initially, only \SI{10}{\percent} of the data are used (\num{292} cut-in scenarios) and the amount of data is gradually increased.
Selection of the samples is done without replacement, so a single data entry is used at most once during a single experiment.
To reduce the impact of randomness, for each selected data size, the experiments are repeated \num{50} times.

\subsection{Answering research question 1}
\label{sec:results rq1}

To answer \cref{rq:exposure}, \cref{fig:cutin llh} shows the mean log-likelihood of the samples that are not used for fitting the \ac{PDF}.
The mean log-likelihood is a measure of how well \iac{PDF} explains observed data.
Mathematically, if $\{\parameters_{\indexremaining}\}_{\indexremaining\in\setindicestest}$ with $\setindicestest$ denotes the set of indices of the scenario parameter vectors that are not used for fitting the \ac{PDF}, then the mean log-likelihood is
\begin{equation}
	\label{eq:mean llh}
	\frac{1}{\cardinality{\setindicestest}} 
	\sum_{\indexremaining\in\setindicestest} \log \density{\parameters}{\parameters_{\indexremaining}}.
\end{equation}
Here, $\cardinality{\setindicestest}$ denotes the cardinality of $\setindicestest$ and equals the number of scenarios that are not used for fitting the \ac{PDF}.
In addition, in \cref{eq:mean llh}, the estimated \ac{PDF} is used for $\densityfunc{\parameters}$.
The solid lines in \cref{fig:cutin llh} show the medians over the \num{50} repetitions, while the colored area denote the \ac{IQR}.
We prefer the median and \ac{IQR} over the mean and standard deviation, as they are less sensitive to the influence of outliers.

\setlength{\figurewidth}{.9\linewidth}
\setlength{\figureheight}{.64\figurewidth}
\begin{figure}
	\centering
\begin{tikzpicture}

\definecolor{darkgray176}{RGB}{176,176,176}
\definecolor{lavender204204255}{RGB}{204,204,255}
\definecolor{lightgray204}{RGB}{204,204,204}
\definecolor{pink255204204}{RGB}{255,204,204}

\begin{axis}[
height=\figureheight,
legend cell align={left},
legend style={
  fill opacity=0.8,
  draw opacity=1,
  text opacity=1,
  at={(0.98,0.02)},
  anchor=south east,
  draw=lightgray204
},
legend style={nodes={scale=0.75, transform shape}},
log basis x={10},
scaled y ticks=false,
tick align=outside,
tick pos=left,
width=\figurewidth,
x grid style={darkgray176},
xlabel={Fraction of data used},
xmin=0.1, xmax=0.923670857187386,
xmode=log,
xtick style={color=black},
xtick={0.1,0.2,0.4,0.6,0.9},
xticklabel style={align=center},
xticklabels={0.1,0.2,0.4,0.6,0.9},
y grid style={darkgray176},
ylabel={Mean log-likelihood},
ymin=-12.2915993928909, ymax=-10.5330671072006,
ytick style={color=black},
yticklabel style={/pgf/number format/fixed,/pgf/number format/precision=3}
]
\path [draw=pink255204204, fill=pink255204204]
(axis cs:0.1,-11.3940795338813)
--(axis cs:0.1,-11.7228805834328)
--(axis cs:0.108263673387405,-11.5952211508711)
--(axis cs:0.117210229753348,-11.6353254711531)
--(axis cs:0.126896100316792,-11.55006718738)
--(axis cs:0.137382379588326,-11.4863399411312)
--(axis cs:0.148735210729351,-11.5640842697327)
--(axis cs:0.161026202756094,-11.3439222225279)
--(axis cs:0.174332882219999,-11.4630268465372)
--(axis cs:0.18873918221351,-11.4266790537771)
--(axis cs:0.204335971785694,-11.3849877742908)
--(axis cs:0.221221629107045,-11.335746309952)
--(axis cs:0.239502661998749,-11.314911322303)
--(axis cs:0.259294379740467,-11.2637023965437)
--(axis cs:0.280721620394118,-11.2305758045473)
--(axis cs:0.30391953823132,-11.2931880687883)
--(axis cs:0.329034456231267,-11.2051192253821)
--(axis cs:0.356224789026244,-11.1638007606844)
--(axis cs:0.385662042116347,-11.1851154620816)
--(axis cs:0.41753189365604,-11.1463619063693)
--(axis cs:0.452035365636024,-11.1125601156984)
--(axis cs:0.489390091847749,-11.1143782790265)
--(axis cs:0.529831690628371,-11.0826859879335)
--(axis cs:0.573615251044868,-11.10765106956)
--(axis cs:0.621016941891562,-11.0455760404137)
--(axis cs:0.672335753649934,-11.0212445230849)
--(axis cs:0.727895384398315,-11.0614799971248)
--(axis cs:0.788046281566991,-10.9920238151924)
--(axis cs:0.853167852417281,-11.0072703637552)
--(axis cs:0.923670857187386,-11.0080087787892)
--(axis cs:0.923670857187386,-10.8318109165639)
--(axis cs:0.923670857187386,-10.8318109165639)
--(axis cs:0.853167852417281,-10.8606147416294)
--(axis cs:0.788046281566991,-10.8760805952946)
--(axis cs:0.727895384398315,-10.9296377227978)
--(axis cs:0.672335753649934,-10.9328070318479)
--(axis cs:0.621016941891562,-10.9463246131648)
--(axis cs:0.573615251044868,-10.9898240921308)
--(axis cs:0.529831690628371,-10.9714398686775)
--(axis cs:0.489390091847749,-10.9879578320605)
--(axis cs:0.452035365636024,-10.9822063250121)
--(axis cs:0.41753189365604,-11.0415174830763)
--(axis cs:0.385662042116347,-11.0601044040288)
--(axis cs:0.356224789026244,-11.064457720222)
--(axis cs:0.329034456231267,-11.0649981686572)
--(axis cs:0.30391953823132,-11.0875363681864)
--(axis cs:0.280721620394118,-11.1265550263766)
--(axis cs:0.259294379740467,-11.1155446632672)
--(axis cs:0.239502661998749,-11.1683319014991)
--(axis cs:0.221221629107045,-11.1741198786588)
--(axis cs:0.204335971785694,-11.2133613859105)
--(axis cs:0.18873918221351,-11.221306841432)
--(axis cs:0.174332882219999,-11.2643692087147)
--(axis cs:0.161026202756094,-11.2203511489299)
--(axis cs:0.148735210729351,-11.2700674355431)
--(axis cs:0.137382379588326,-11.2632746982729)
--(axis cs:0.126896100316792,-11.3020186841498)
--(axis cs:0.117210229753348,-11.3433754536975)
--(axis cs:0.108263673387405,-11.3781334839874)
--(axis cs:0.1,-11.3940795338813)
--cycle;
\addlegendimage{area legend, draw=pink255204204, fill=pink255204204}
\addlegendentry{KDE, IQR}

\path [draw=lavender204204255, fill=lavender204204255]
(axis cs:0.1,-11.3327438831329)
--(axis cs:0.1,-11.8824365139008)
--(axis cs:0.108263673387405,-12.2116661071777)
--(axis cs:0.117210229753348,-11.7217328548431)
--(axis cs:0.126896100316792,-11.802806854248)
--(axis cs:0.137382379588326,-11.5591244697571)
--(axis cs:0.148735210729351,-11.4200024604797)
--(axis cs:0.161026202756094,-11.4796650409698)
--(axis cs:0.174332882219999,-11.4414746761322)
--(axis cs:0.18873918221351,-11.3917486667633)
--(axis cs:0.204335971785694,-11.4529070854187)
--(axis cs:0.221221629107045,-11.2298276424408)
--(axis cs:0.239502661998749,-11.1666162014008)
--(axis cs:0.259294379740467,-11.1317856311798)
--(axis cs:0.280721620394118,-11.0542986392975)
--(axis cs:0.30391953823132,-11.1107597351074)
--(axis cs:0.329034456231267,-10.9533305168152)
--(axis cs:0.356224789026244,-10.9755702018738)
--(axis cs:0.385662042116347,-10.9561724662781)
--(axis cs:0.41753189365604,-10.909526348114)
--(axis cs:0.452035365636024,-10.843519449234)
--(axis cs:0.489390091847749,-10.9327867031097)
--(axis cs:0.529831690628371,-10.8703618049622)
--(axis cs:0.573615251044868,-10.8373553752899)
--(axis cs:0.621016941891562,-10.8318531513214)
--(axis cs:0.672335753649934,-10.818727016449)
--(axis cs:0.727895384398315,-10.7991383075714)
--(axis cs:0.788046281566991,-10.7527775764465)
--(axis cs:0.853167852417281,-10.8010654449463)
--(axis cs:0.923670857187386,-10.835066318512)
--(axis cs:0.923670857187386,-10.6130003929138)
--(axis cs:0.923670857187386,-10.6130003929138)
--(axis cs:0.853167852417281,-10.6386411190033)
--(axis cs:0.788046281566991,-10.647894859314)
--(axis cs:0.727895384398315,-10.6647021770477)
--(axis cs:0.672335753649934,-10.7033586502075)
--(axis cs:0.621016941891562,-10.7082192897797)
--(axis cs:0.573615251044868,-10.733672618866)
--(axis cs:0.529831690628371,-10.7478973865509)
--(axis cs:0.489390091847749,-10.7890672683716)
--(axis cs:0.452035365636024,-10.7356276512146)
--(axis cs:0.41753189365604,-10.8066489696503)
--(axis cs:0.385662042116347,-10.8102679252625)
--(axis cs:0.356224789026244,-10.8210282325745)
--(axis cs:0.329034456231267,-10.8342437744141)
--(axis cs:0.30391953823132,-10.8979506492615)
--(axis cs:0.280721620394118,-10.8821671009064)
--(axis cs:0.259294379740467,-10.8964231014252)
--(axis cs:0.239502661998749,-10.9043490886688)
--(axis cs:0.221221629107045,-11.0159404277802)
--(axis cs:0.204335971785694,-11.0459139347076)
--(axis cs:0.18873918221351,-11.0240590572357)
--(axis cs:0.174332882219999,-11.0205707550049)
--(axis cs:0.161026202756094,-11.0775909423828)
--(axis cs:0.148735210729351,-11.0975458621979)
--(axis cs:0.137382379588326,-11.1059727668762)
--(axis cs:0.126896100316792,-11.1671130657196)
--(axis cs:0.117210229753348,-11.2343599796295)
--(axis cs:0.108263673387405,-11.1968398094177)
--(axis cs:0.1,-11.3327438831329)
--cycle;
\addlegendimage{area legend, draw=lavender204204255, fill=lavender204204255}
\addlegendentry{NF, IQR}

\addplot [semithick, red]
table {%
0.1 -11.4898261004594
0.108263673387405 -11.4729600749679
0.117210229753348 -11.4072812765405
0.126896100316792 -11.3844562828213
0.137382379588326 -11.3463833413978
0.148735210729351 -11.3602030190337
0.161026202756094 -11.2869351021097
0.174332882219999 -11.3392870385752
0.18873918221351 -11.3238140620537
0.204335971785694 -11.2846279287797
0.221221629107045 -11.2545747684653
0.239502661998749 -11.2368985079643
0.259294379740467 -11.1725798828918
0.280721620394118 -11.1891343857548
0.30391953823132 -11.1949293152606
0.329034456231267 -11.1220790905291
0.356224789026244 -11.1085725333591
0.385662042116347 -11.1201235174418
0.41753189365604 -11.0890211810075
0.452035365636024 -11.0446541626166
0.489390091847749 -11.0427144867546
0.529831690628371 -11.0221018115436
0.573615251044868 -11.0339692498972
0.621016941891562 -10.9918915827082
0.672335753649934 -10.9800064323439
0.727895384398315 -10.9828215102892
0.788046281566991 -10.9292180884677
0.853167852417281 -10.9309178971873
0.923670857187386 -10.9163255521774
};
\addlegendentry{KDE, median}
\addplot [semithick, blue]
table {%
0.1 -11.5404887199402
0.108263673387405 -11.4006357192993
0.117210229753348 -11.3772706985474
0.126896100316792 -11.3439431190491
0.137382379588326 -11.2476425170898
0.148735210729351 -11.2326803207397
0.161026202756094 -11.1757855415344
0.174332882219999 -11.1699247360229
0.18873918221351 -11.1421675682068
0.204335971785694 -11.125675201416
0.221221629107045 -11.1185126304626
0.239502661998749 -10.95924949646
0.259294379740467 -10.9829678535461
0.280721620394118 -10.9664497375488
0.30391953823132 -10.9581789970398
0.329034456231267 -10.8608660697937
0.356224789026244 -10.8976349830627
0.385662042116347 -10.8717217445374
0.41753189365604 -10.8547897338867
0.452035365636024 -10.7771983146667
0.489390091847749 -10.8396544456482
0.529831690628371 -10.795015335083
0.573615251044868 -10.7846994400024
0.621016941891562 -10.7609090805054
0.672335753649934 -10.748354434967
0.727895384398315 -10.7316484451294
0.788046281566991 -10.710391998291
0.853167852417281 -10.7263026237488
0.923670857187386 -10.7037825584412
};
\addlegendentry{NF, median}
\end{axis}

\end{tikzpicture}
	\caption{Mean log-likelihood of the samples that are not used for fittings the \ac{PDF}.
		The solid lines show the median of the \num{50} repetitions, while the colored areas denote the \acf{IQR}.}
	\label{fig:cutin llh}
\end{figure}
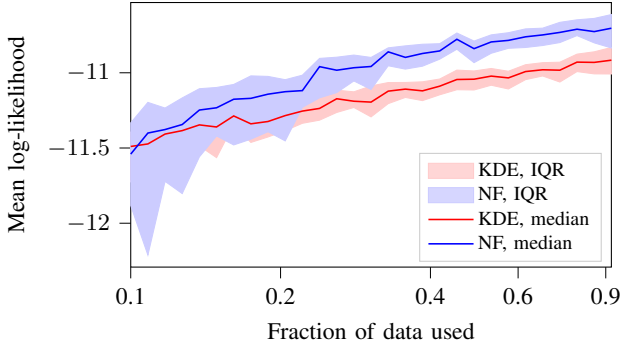

\Cref{fig:cutin llh} shows that both \ac{NF} and \ac{KDE} perform better if more data are used.
This result can be expected, as the use of more data generally results in better fits.
\Cref{fig:cutin llh} also shows that \ac{NF} mostly outperforms \ac{KDE}. 
Only when using \SI{10}{\percent} of the data, \ac{KDE} scores slightly better than \ac{NF} when considering the median.
When using more than \SI{30}{\percent} of the data, the \ac{IQR} of \ac{NF} is entirely above the \ac{IQR} of \ac{KDE}, indicating that \ac{NF} provide a significantly better fit of the \ac{PDF}.

For quantifying the risk, a good approximation $\densityisfunc{\parameters}$ of the entire domain $\realnumbers^{\dimension}$ is not important.
Rather, as can be seen in \cref{eq:risk}, all that matters is a good approximation of $\densityisfunc{\parameters}$ for the values of $\parameters$ where $\simulationoutcome{\parameters}=1$.
For a good \ac{ADS}, it can be expected that collisions are rare, such that $\simulationoutcome{\parameters}=1$ only when $\parameters$ is near the boundary.
Hence, we are interested in how \ac{NF} and \ac{KDE} compare in estimating the \ac{PDF} of the scenario parameters near the boundaries.
For this, we have evaluated the mean log-likelihood of \cref{eq:mean llh} with $\setindicestest$ replaced by $\setindicaspareto$, where $\setindicaspareto$ denotes the indices of the scenarios that are at the Pareto front of the remaining scenarios.
Thus,
\begin{align*}
	\setindicaspareto =& \left\{ 
		\indexremaining\in\setindicestest : \{ 
			\indexremaining'\in\setindicestest: 
			\parameters_{\indexremaining'} \succ \parameters_{\indexremaining}
		\} = \emptyset
	\right\} \,\cup \\
	&\quad \left\{ 
		\indexremaining\in\setindicestest : \{ 
			\indexremaining'\in\setindicestest: 
			\parameters_{\indexremaining} \succ \parameters_{\indexremaining'}
		\} = \emptyset
	\right\}
\end{align*}
where $\parameters_{\indexremaining'} \succ \parameters_{\indexremaining}$ indicates that all $\dimension$ entries of $\parameters_{\indexremaining'}$ are strictly larger than the corresponding entries of $\parameters_{\indexremaining}$.

\begin{figure}
	\centering
\begin{tikzpicture}

\definecolor{darkgray176}{RGB}{176,176,176}
\definecolor{lavender204204255}{RGB}{204,204,255}
\definecolor{lightgray204}{RGB}{204,204,204}
\definecolor{pink255204204}{RGB}{255,204,204}

\begin{axis}[
height=\figureheight,
legend cell align={left},
legend style={
  fill opacity=0.8,
  draw opacity=1,
  text opacity=1,
  at={(0.98,0.02)},
  anchor=south east,
  draw=lightgray204
},
legend style={nodes={scale=0.75, transform shape}},
log basis x={10},
scaled y ticks=false,
tick align=outside,
tick pos=left,
width=\figurewidth,
x grid style={darkgray176},
xlabel={Fraction of data used},
xmin=0.1, xmax=0.923670857187386,
xmode=log,
xtick style={color=black},
xtick={0.1,0.2,0.4,0.6,0.9},
xticklabel style={align=center},
xticklabels={0.1,0.2,0.4,0.6,0.9},
y grid style={darkgray176},
ylabel={Mean log-likelihood},
ymin=-19, ymax=-12,
ytick style={color=black},
yticklabel style={/pgf/number format/fixed,/pgf/number format/precision=3}
]
\path [draw=pink255204204, fill=pink255204204]
(axis cs:0.1,-14.7192459667392)
--(axis cs:0.1,-16.9539716514946)
--(axis cs:0.108263673387405,-17.4040856918165)
--(axis cs:0.117210229753348,-16.5654816426154)
--(axis cs:0.126896100316792,-16.3339819185424)
--(axis cs:0.137382379588326,-15.7681891101455)
--(axis cs:0.148735210729351,-16.089246369151)
--(axis cs:0.161026202756094,-15.8473744104366)
--(axis cs:0.174332882219999,-15.6614800651317)
--(axis cs:0.18873918221351,-15.9330265442304)
--(axis cs:0.204335971785694,-15.3858490171426)
--(axis cs:0.221221629107045,-15.424566735522)
--(axis cs:0.239502661998749,-15.3191150502039)
--(axis cs:0.259294379740467,-15.1558076864308)
--(axis cs:0.280721620394118,-15.1540921147402)
--(axis cs:0.30391953823132,-14.9446811669912)
--(axis cs:0.329034456231267,-14.8301228689066)
--(axis cs:0.356224789026244,-15.1053702507992)
--(axis cs:0.385662042116347,-15.132118775553)
--(axis cs:0.41753189365604,-14.9332397042133)
--(axis cs:0.452035365636024,-14.6931754363522)
--(axis cs:0.489390091847749,-14.6551861744645)
--(axis cs:0.529831690628371,-14.2968860922398)
--(axis cs:0.573615251044868,-14.5750937607542)
--(axis cs:0.621016941891562,-14.1641069549608)
--(axis cs:0.672335753649934,-14.2271152746534)
--(axis cs:0.727895384398315,-13.919218339139)
--(axis cs:0.788046281566991,-13.506906206349)
--(axis cs:0.853167852417281,-13.2911799218615)
--(axis cs:0.923670857187386,-12.779137701698)
--(axis cs:0.923670857187386,-12.2194064786618)
--(axis cs:0.923670857187386,-12.2194064786618)
--(axis cs:0.853167852417281,-12.6105045704497)
--(axis cs:0.788046281566991,-13.010879210714)
--(axis cs:0.727895384398315,-13.4176354702103)
--(axis cs:0.672335753649934,-13.4275451750909)
--(axis cs:0.621016941891562,-13.4618404003217)
--(axis cs:0.573615251044868,-13.7263944503793)
--(axis cs:0.529831690628371,-13.6735329950384)
--(axis cs:0.489390091847749,-14.0425877171074)
--(axis cs:0.452035365636024,-13.9862120677591)
--(axis cs:0.41753189365604,-14.0683743848282)
--(axis cs:0.385662042116347,-14.0619753194277)
--(axis cs:0.356224789026244,-14.2601412444597)
--(axis cs:0.329034456231267,-14.1377471959559)
--(axis cs:0.30391953823132,-14.1799773588447)
--(axis cs:0.280721620394118,-14.3457191597374)
--(axis cs:0.259294379740467,-14.3344967121456)
--(axis cs:0.239502661998749,-14.2386810994047)
--(axis cs:0.221221629107045,-14.307426909682)
--(axis cs:0.204335971785694,-14.4074633382966)
--(axis cs:0.18873918221351,-14.4010707424893)
--(axis cs:0.174332882219999,-14.3907212615463)
--(axis cs:0.161026202756094,-14.6060199034402)
--(axis cs:0.148735210729351,-14.6390970882927)
--(axis cs:0.137382379588326,-14.5945686503697)
--(axis cs:0.126896100316792,-14.6788649310636)
--(axis cs:0.117210229753348,-14.6274758040208)
--(axis cs:0.108263673387405,-14.8024974991735)
--(axis cs:0.1,-14.7192459667392)
--cycle;
\addlegendimage{area legend, draw=pink255204204, fill=pink255204204}
\addlegendentry{KDE, IQR}

\path [draw=lavender204204255, fill=lavender204204255]
(axis cs:0.1,-16.831157207489)
--(axis cs:0.1,-24.6665263175964)
--(axis cs:0.108263673387405,-26.0840048789978)
--(axis cs:0.117210229753348,-26.0358939170837)
--(axis cs:0.126896100316792,-24.4629573822021)
--(axis cs:0.137382379588326,-20.1269302368164)
--(axis cs:0.148735210729351,-20.9074091911316)
--(axis cs:0.161026202756094,-18.4240279197693)
--(axis cs:0.174332882219999,-19.1840400695801)
--(axis cs:0.18873918221351,-18.5447764396667)
--(axis cs:0.204335971785694,-18.0639867782593)
--(axis cs:0.221221629107045,-16.4851188659668)
--(axis cs:0.239502661998749,-17.1696538925171)
--(axis cs:0.259294379740467,-15.8870315551758)
--(axis cs:0.280721620394118,-15.7436349391937)
--(axis cs:0.30391953823132,-15.4227714538574)
--(axis cs:0.329034456231267,-15.2146787643433)
--(axis cs:0.356224789026244,-15.5767142772675)
--(axis cs:0.385662042116347,-15.0705316066742)
--(axis cs:0.41753189365604,-14.9777636528015)
--(axis cs:0.452035365636024,-14.7626144886017)
--(axis cs:0.489390091847749,-15.0593612194061)
--(axis cs:0.529831690628371,-14.396317243576)
--(axis cs:0.573615251044868,-14.1771621704102)
--(axis cs:0.621016941891562,-13.9540960788727)
--(axis cs:0.672335753649934,-13.9915342330933)
--(axis cs:0.727895384398315,-13.7831637859344)
--(axis cs:0.788046281566991,-13.5014910697937)
--(axis cs:0.853167852417281,-13.0609769821167)
--(axis cs:0.923670857187386,-12.7155733108521)
--(axis cs:0.923670857187386,-12.0247464179993)
--(axis cs:0.923670857187386,-12.0247464179993)
--(axis cs:0.853167852417281,-12.4363451004028)
--(axis cs:0.788046281566991,-12.9719650745392)
--(axis cs:0.727895384398315,-13.2795295715332)
--(axis cs:0.672335753649934,-13.2715110778809)
--(axis cs:0.621016941891562,-13.3237776756287)
--(axis cs:0.573615251044868,-13.7192118167877)
--(axis cs:0.529831690628371,-13.727888584137)
--(axis cs:0.489390091847749,-14.0626962184906)
--(axis cs:0.452035365636024,-14.095903635025)
--(axis cs:0.41753189365604,-14.1838762760162)
--(axis cs:0.385662042116347,-14.3164224624634)
--(axis cs:0.356224789026244,-14.6004512310028)
--(axis cs:0.329034456231267,-14.4460599422455)
--(axis cs:0.30391953823132,-14.7519779205322)
--(axis cs:0.280721620394118,-14.6723477840424)
--(axis cs:0.259294379740467,-14.8037672042847)
--(axis cs:0.239502661998749,-14.7916493415833)
--(axis cs:0.221221629107045,-14.9705283641815)
--(axis cs:0.204335971785694,-15.2356405258179)
--(axis cs:0.18873918221351,-15.1475429534912)
--(axis cs:0.174332882219999,-15.2003605365753)
--(axis cs:0.161026202756094,-15.4503283500671)
--(axis cs:0.148735210729351,-16.0694942474365)
--(axis cs:0.137382379588326,-15.6393795013428)
--(axis cs:0.126896100316792,-16.1974964141846)
--(axis cs:0.117210229753348,-16.1106014251709)
--(axis cs:0.108263673387405,-16.355929851532)
--(axis cs:0.1,-16.831157207489)
--cycle;
\addlegendimage{area legend, draw=lavender204204255, fill=lavender204204255}
\addlegendentry{NF, IQR}

\addplot [semithick, red]
table {%
0.1 -15.7008429961089
0.108263673387405 -15.8167170320786
0.117210229753348 -15.4455172602328
0.126896100316792 -15.3911856064491
0.137382379588326 -15.0490210543433
0.148735210729351 -15.2935975612283
0.161026202756094 -15.094829575636
0.174332882219999 -14.9206164208758
0.18873918221351 -14.919491081101
0.204335971785694 -14.9123828454956
0.221221629107045 -14.6348184298235
0.239502661998749 -14.6623456319002
0.259294379740467 -14.6988704674947
0.280721620394118 -14.7945960643363
0.30391953823132 -14.5224433412892
0.329034456231267 -14.5408363312138
0.356224789026244 -14.7833718552536
0.385662042116347 -14.5051237846006
0.41753189365604 -14.3440383800476
0.452035365636024 -14.248069048473
0.489390091847749 -14.3015998131398
0.529831690628371 -13.887495861192
0.573615251044868 -13.9617904809718
0.621016941891562 -13.8405388314698
0.672335753649934 -13.7242692364261
0.727895384398315 -13.6825465452613
0.788046281566991 -13.1409998260123
0.853167852417281 -12.9899696015184
0.923670857187386 -12.5039765580299
};
\addlegendentry{KDE, median}
\addplot [semithick, blue]
table {%
0.1 -19.3770799636841
0.108263673387405 -18.646785736084
0.117210229753348 -18.0452556610107
0.126896100316792 -17.1585845947266
0.137382379588326 -16.4751377105713
0.148735210729351 -16.8548622131348
0.161026202756094 -16.2205257415771
0.174332882219999 -15.9493680000305
0.18873918221351 -15.7208385467529
0.204335971785694 -15.9303002357483
0.221221629107045 -15.4142370223999
0.239502661998749 -15.2074933052063
0.259294379740467 -15.2647953033447
0.280721620394118 -15.142110824585
0.30391953823132 -14.9635310173035
0.329034456231267 -14.7213373184204
0.356224789026244 -14.7881646156311
0.385662042116347 -14.7374911308289
0.41753189365604 -14.5514364242554
0.452035365636024 -14.3511972427368
0.489390091847749 -14.41157579422
0.529831690628371 -14.0028867721558
0.573615251044868 -13.9306888580322
0.621016941891562 -13.5995111465454
0.672335753649934 -13.6063623428345
0.727895384398315 -13.5021681785583
0.788046281566991 -13.1652770042419
0.853167852417281 -12.7713074684143
0.923670857187386 -12.4137978553772
};
\addlegendentry{NF, median}
\end{axis}

\end{tikzpicture}
	\caption{Similar to \cref{fig:cutin llh}, but now only the samples at the Pareto front are considered for calculating the mean log-likelihood.}
	\label{fig:cutin llh pareto}
\end{figure}
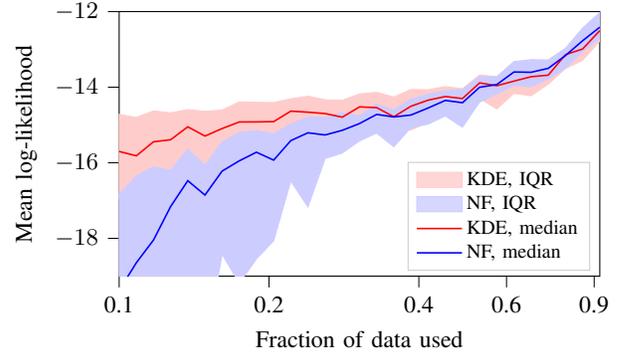

\Cref{fig:cutin llh pareto} shows the mean log-likelihoods for the \ac{NF} and \ac{KDE} when only considering the scenario parameters that are at the Pareto front.
Whereas \ac{NF} outperforms \ac{KDE} in \cref{fig:cutin llh}, now \ac{KDE} generally provides equal or better scores. 
Especially when using \SI{40}{\percent} or less data, \ac{KDE} provides higher likelihoods, which is significant according to the Mann-Whitney U test. 
When using more that \SI{40}{\percent}, \ac{NF} and \ac{KDE} perform approximately similar.

In conclusion, \ac{NF} perform significantly better in estimating the \ac{PDF} of the scenario parameters compared to \ac{KDE} (see \cref{fig:cutin llh}).
However, when considering the extreme values of the scenario parameters, \ac{KDE} performs significantly better than \ac{NF} if only \SI{40}{\percent} or less data is used, while with more data, \ac{KDE} and \ac{NF} perform similar (see \cref{fig:cutin llh pareto}).

\subsection{Answering research question 2}
\label{sec:results rq2}

\begin{figure}
	\centering
	\input{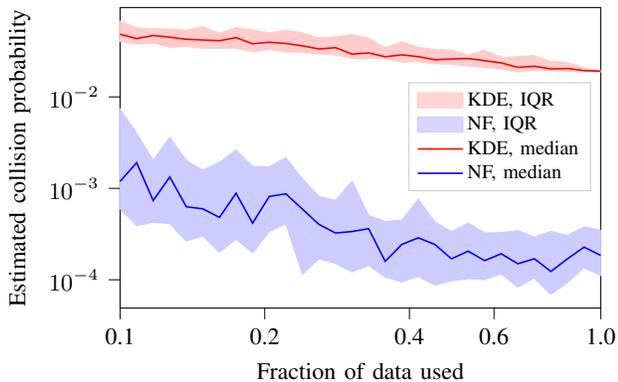}
	\caption{Estimated collision risk using \cref{eq:is}.
		The solid lines show the median of the \num{50} repetitions, while the colored areas denote the \acf{IQR}.}
	\label{fig:cutin iqr}
\end{figure}

\Cref{fig:cutin iqr} shows the estimated collision probability using \cref{eq:is} with $\numberofis=\numberofmc=\num{10000}$ while using different amounts of data. 
Clearly, the estimated collision probability is substantially higher if \ac{KDE} is used compared to \ac{NF}.
On average, with \ac{KDE}, the estimated collision probability is roughly \num{40} times higher (note the log scale in \cref{fig:cutin iqr}).
As no ground truth is available, we cannot argue that one approach is better.
It might be argued that \ac{NF} assigns too little probability mass to the tails of the distribution \autocite{hickling2024flexible}.
However, especially when using at least \SI{40}{\percent} of the data, \ac{NF} do not perform worse in terms of the mean log-likelihood of the extreme values as we have seen in \cref{fig:cutin llh pareto}.
In addition, it might be argued that \ac{KDE} overestimates the probability density near the tails of the distribution \autocite{beranger2019tail}.

\setlength{\figureheight}{.7\figurewidth}

\Cref{fig:cutin iqr} shows that the estimated collision probability decreases with the use of more data.
For \ac{KDE}, this effect has already been explained in \autocite{degelder2023certain}: with less data, the bandwidth used for the \ac{KDE} tends to be larger, leading to larger tails of the estimated \ac{PDF}.
Since the collisions occur at the tails of the distribution, this also leads to an overestimation. 
Although \ac{NF} assign substantially less mass to the probability density near the tails, resulting in a lower estimated probability collision, a similar effect can be observed.

The \acp{IQR} in \cref{fig:cutin iqr} show that the variance of the results with \ac{NF} is relatively larger compared to \ac{KDE}.
In absolute terms, however, the \ac{IQR} for \ac{KDE} is larger (note the log scale). 
Both results are not unexpected: when estimating a smaller probability, the uncertainty scales roughly with the square root of this probability, meaning that the \ac{IQR} will be smaller with smaller medians and that the ratio of \ac{IQR} and the median will increase with smaller estimations.
However, when a large part of the data is used, the \ac{IQR} with \ac{KDE} gets significantly smaller.
When using all data, the \ac{IQR} with \ac{KDE} is almost zero.
This can be explained from the fact that when using all data, this results in \num{50} times the same result \ac{PDF} estimation with \ac{KDE}, since \ac{KDE} is fully deterministic.
Thus, the only difference in the \num{50} repetitions comes from the crude Monte Carlo and importance sampling, which only contribute little to the overall variance, as already observed in \autocite{degelder2023certain}.
For \ac{NF}, however, the \ac{PDF} estimations still vary due to the random initializations of the \ac{NF}, as during the \num{50} repetitions, a different seed has been used for initializing the \ac{NF} before training.

In conclusion, the results of the estimated collision probability differ substantially when using \ac{NF} or \ac{KDE} (see \cref{fig:cutin iqr}). 
When using \ac{NF}, the estimated collision probability is an order of magnitude lower.
An explanation for this outcome is that \ac{KDE} generally assigns more probability mass to the tails of the \ac{PDF} estimation and the collisions occur at these tails of the \ac{PDF}.

	\section{DISCUSSION}
\label{sec:discussion}

This paper's objective was to compare \ac{NF} with \ac{KDE} in estimating the \ac{PDF} of scenario parameters and compare the estimated \ac{ADS} risk with these estimated \acp{PDF}.
The results show that \ac{NF} generally provide significantly better estimations of the \ac{PDF}, while at the extremes of the data, the differences are more subtle with \ac{KDE} performing better than \ac{NF} with low amounts of training data and equal performance otherwise.
When estimating the risk, it appears that the estimated risk is substantially higher when \ac{KDE} is used.
A possible explanation is that \ac{KDE} overestimates the probability density near the tails of the distribution \autocite{beranger2019tail} and as a result, a higher collision probability is estimated.
Besides these findings, there may be other reasons to prefer \ac{NF} over \ac{KDE} or vice versa, which are further discussed in this section.

As discussed in \cref{sec:risk quantification}, for determining the importance density $\densityisfunc{\parameters}$, \ac{KDE} has been used, also if \ac{NF} has been used for estimating $\densityfunc{\parameters}$. 
The reason for this is that we prefer an importance density that is larger than the ideal importance density at the tails, rather than an importance density for which $\densityisfunc{\parameters} \ll \simulationoutcome{\parameters}\density{\parameters}{\parameters}$.
As can be seen from \cref{eq:is}, if the latter is the case, this would either lead to a high variance in the outcome, or in a biased outcome since we would not sample values of $\parameters$ where $\densityisfunc{\parameters} \ll \simulationoutcome{\parameters}\density{\parameters}{\parameters}$.
Given that \ac{KDE} generally overestimates the tails, it might result in a slower convergence, but we compensated this with a rather large number of simulations.
Note that with longer simulation times or higher dimensional scenario parameter vectors, \ac{NF} might be preferred to estimate $\densityisfunc{\parameters}$.

In general, \ac{KDE} suffer from the curse of dimensionality \autocite{scott2015multivariate}, while \ac{NF} scale better with higher dimensions \autocite{rezende2015variational}.
Thus, when providing higher-dimensional data, \ac{NF} may be preferred.
In this study, we have applied \ac{KDE} and \ac{NF} on $\dimension=4$-dimensional data.
Therefore, we have not explored the potential of \ac{NF} with higher-dimensional data.
In a future publication, we will study the use of \ac{NF} for the assessment of \acp{ADS} when using higher-dimensional data, which enables to capture more details of the scenarios with the scenario parameters.

Another consideration is computational speed. 
For fitting the \ac{PDF}, \ac{KDE} is generally faster than \ac{NF}, which require substantial training time. 
Computational speed was not the primary focus of this study, so only limited conclusions can be drawn. 
However, fitting the \ac{KDE} on the full dataset took approximately \SI{1.6}{\second}, whereas with \ac{NF}, it required around \SI{95}{\second}, making it roughly \num{60} times slower.
Even with complex bandwidths, such as a full-rank bandwidth matrix, \ac{KDE} is likely to remain more efficient.
However, the computational speed for evaluating the \ac{PDF} varies. 
For \ac{NF}, it depends on model complexity, while for \ac{KDE}, it depends on the number of samples used. 
With small datasets, \ac{KDE} may evaluate densities faster than \ac{NF}, but this advantage diminishes as the dataset size increases.

The last consideration we will discuss is ease of use.
\ac{KDE} is more straightforward to implement.
While \ac{NF} may potentially offer superior \ac{PDF} estimation, their performance depends on various choices, including the choice of architecture (e.g., number and size of layers), types of transformations, and optimization hyperparameters. 
Therefore, if ease of use is a priority, \ac{KDE} is the preferable option.

	
    \acresetall
	\section{CONCLUSIONS}
\label{sec:conclusions}

This study compared \ac{NF} and \ac{KDE} in the context of risk quantification for \acp{ADS}. 
Our findings show that \ac{NF} outperform \ac{KDE} in estimating the \ac{PDF} of the scenario parameters.
\Ac{KDE}, while simpler and computationally more efficient to train, tends to overestimate probability density in the tails, which can lead to higher risk estimations. 

Future work will explore the application of \ac{NF} for scenario generation, particularly for modeling more complex scenarios involving a larger number of parameters. 
Additionally, we aim to investigate how different \ac{NF} architectures, transformation strategies, and training procedures affect performance in terms of accuracy and computational cost. 
Another topic for future research is the comparison of \ac{NF} and \ac{KDE} with other density estimators like variational autoencoders \autocite{doersch2016tutorial} and diffusion models \autocite{premkumar2024diffusion}.
In this way, we seek to improve the overall robustness of scenario-based safety assessment frameworks for \acp{ADS}.

	\printbibliography

\end{document}